\pdfoutput=1 

\documentclass[10pt,twocolumn,letterpaper]{article}

\usepackage{cvpr}              
\makeatletter
\@namedef{ver@everyshi.sty}{}
\makeatother

\usepackage{graphicx}
\usepackage{amsmath}
\usepackage{amssymb}
\usepackage{booktabs}
\usepackage{xcolor}
\usepackage{array,multirow,textcomp}
\usepackage{makecell}
\usepackage{csquotes}
\usepackage{epigraph}
\usepackage{adjustbox}
\usepackage{enumitem}
\usepackage{tikz}

\newcommand*\circled[1]{%
  \tikz[baseline=(C.base)]\node[draw,circle,inner sep=1.2pt,line width=0.2mm,](C) {#1};}
\newcommand*\Myitem{%
  \stepcounter{enumi}\item[\circled{\theenumi}]}

\usepackage[pagebackref,breaklinks,colorlinks]{hyperref}

\usepackage[capitalize]{cleveref}
\crefname{section}{Sec.}{Secs.}
\Crefname{section}{Section}{Sections}
\Crefname{table}{Table}{Tables}
\crefname{table}{Tab.}{Tabs.}

\definecolor{mygray}{gray}{.9}

\newcommand{\greenA}[0]{\color{green} \uparrow}
\newcommand{\redA}[0]{\color{red} \downarrow}

\newcommand{\tableFont}{\fontsize{5.5pt}{6.6pt} \selectfont}

\newcommand{\trendCite}{~\cite{Rakin2019ParametricNI,Kundu2021HIRE_SNN,Eustratiadis2021WeightCovAlign,Lecuyer2019CertifiedDifferential,Yang2022FeatureUncrt,Jeddi_2020_CVPR,Addepalli_2021_CVPR,Lee2021GradDivAR}}
\newcommand{\trendCiteShort}{~\cite{Rakin2019ParametricNI,Kundu2021HIRE_SNN,Eustratiadis2021WeightCovAlign}}

\def\cvprPaperID{13} 
\def\confName{CVPR}
\def\confYear{2022}

\begin{document}


\title{Gradient Obfuscation Checklist Test Gives a False Sense of Security}

\author{
Nikola Popovic$^{1}$, Danda Pani Paudel$^{1}$, Thomas Probst$^{1}$, Luc Van Gool$^{1,2}$\\
$^{1}$Computer Vision Laboratory, ETH Zurich, Switzerland\\
$^{2}$VISICS, ESAT/PSI, KU Leuven, Belgium\\
{\tt\small \{nipopovic, paudel, probstt, vangool\}@vision.ee.ethz.ch}
}

\maketitle

\begin{abstract}

One popular group of defense techniques against adversarial attacks is based on injecting stochastic noise into the network. The main source of robustness of such stochastic defenses however is often due to the obfuscation of the gradients, offering a false sense of security.
Since most of the popular adversarial attacks are optimization-based, obfuscated gradients reduce their attacking ability, while the model is still susceptible to stronger or specifically tailored adversarial attacks.
Recently, five characteristics have been identified, which are commonly observed when the improvement in robustness is mainly caused by gradient obfuscation.
It has since become a trend to use these five characteristics as a sufficient test, to determine whether or not gradient obfuscation is the main source of robustness.
However, these characteristics do not perfectly characterize all existing cases of gradient obfuscation, and therefore can not serve as a basis for a conclusive test.
In this work, we present a counterexample, showing this test is not sufficient for concluding that gradient obfuscation is not the main cause of improvements in robustness.

\end{abstract}
\section{Introduction}
\label{sec:introduction}

Deep Neural Networks (DNN) achieved astonishing results in the last decade, resulting in breakthroughs in processing images, videos, speech, audio, and natural language~\cite{lecun2015deep}. 
These networks have the potential to serve as the core of solutions to many real-world problems.
However, it was discovered that a small adversarial perturbation in pixel intensities can cause a severe drop in the performance of DNNs, or worse, make them give a specific false prediction desired by the adversary~\cite{Szegedy2014IntriguingPO,Goodfellow2015ExplainingAH}.
This can have severe consequences in applications like autonomous vehicles, healthcare, etc.
Therefore, it is very important to design defense mechanisms to make DNNs more robust to adversarial attacks, as well as to thoroughly understand the vulnerabilities of a certain model to these attacks.


Numerous approaches have been proposed to defend against adversarial attacks.
Some of the main defense categories are
adversarial training~\cite{Kurakin2017AdversarialML,Wong2020FastIB,Madry2018TowardsDL,Zhang2019TheoreticallyPT} 
, certified robustness~\cite{Cohen2019CertifiedAR,Croce2019ProvableRO,Salman2019ProvablyRD} 
and gradient regularization~\cite{Ciss2017ParsevalNI,Ross2018ImprovingTA,Jakubovitz2018ImprovingDR,Gu2015TowardsDN,Finlay2019ScaleableIG}. 
Another popular category of adversarial defenses is noise injection~\cite{Rakin2019ParametricNI,Kundu2021HIRE_SNN,Eustratiadis2021WeightCovAlign,Lecuyer2019CertifiedDifferential}, where some form of stochastic noise is introduced, in an attempt to increase robustness.

Most of the popular adversarial attack methods exploit the network's differentiability to craft the sought adversarial examples~\cite{Szegedy2014IntriguingPO,Goodfellow2015ExplainingAH,Madry2018TowardsDL,Athalye2018SynthesizingRA,mihajlovic2018CommonAdv}. 
Introducing stochastic noise usually weakens these attacks by obfuscating the gradients, creating only apparent robustness.
\emph{Athalye et al.}~\cite{Athalye2018ObfuscatedGG} showed that Expectation over Transformation (EoT), a simple method for gradient estimation, suffices to unveil the obfuscated gradients in those scenarios.
In other words, after using the EoT gradient estimation, many defenses become ineffective.
Furthermore, five characteristics have been identified by \emph{Athalye et al.}, which commonly occur when the improvement in robustness is mainly caused by gradient obfuscation. 
It has since become a trend to use these five characteristics as a sufficient test, to determine whether or not gradient obfuscation is the main source of robustness. 
We empirically show by a counterexample that these characteristics do not characterize all existing cases of gradient obfuscation. Therefore, we argue that the gradient obfuscation checklist test gives a false sense of security.
\section{Detecting Gradient Obfuscation}
\noindent We list the five common characteristics, as observed by \mbox{\emph{Athalye et al.}~\cite{Athalye2018ObfuscatedGG}}, in the following.
\begin{enumerate}[font=\itshape]
    \Myitem One-step attacks perform better than iterative attacks.
    \Myitem Black-box attacks are better than white-box attacks.
    \Myitem Unbounded attacks do not reach $100\%$ success.
    \Myitem Random sampling finds adversarial examples.
    \Myitem Increasing distortion bound does not increase success.
\end{enumerate}
In fact, \emph{Athalye et al.} also mentioned that the above list may not perfectly characterize all the cases of gradient obfuscation.
Despite that, it has recently become a trend to use these five characteristics as criteria of a \mbox{``checklist``}, to determine whether or not the success of a stochastic defense is mainly caused by obfuscating the gradients\trendCite.
As a result, any given defense is claimed
to provide the robustness beyond gradient obfuscation, if none of these five characteristics is observed.

In this work, we empirically show that such a claim can not be made.
Our empirical study unveils a counterexample to the claim.
In particular, we show that the Parametric Noise Injection (PNI) defense~\cite{Rakin2019ParametricNI}, which does not exhibit any of the five characteristics, is still vulnerable to attacks with the EoT gradient estimation. Therefore, its improvement in robustness is mostly based on the obfuscation of gradients. This indicates that the five characteristics are insufficient to be used to determine the contribution of gradient obfuscation to robustness, in general.






\section{Parametric Noise Injection (PNI)}
\label{sec:PNI}

In this section we give an overview of the PNI~\cite{Rakin2019ParametricNI} adversarial defense technique. 

This method injects noise to different components or location within the DNN in the following way:

\begin{equation}
    \label{eq:PNI}
    \begin{aligned}
    &\Tilde{v}_i = f_{\text{PNI}}(v_i) = v_i + \alpha_i \cdot \eta, \\ &\eta \sim \mathcal{N}(0,\sigma^2), \\
    &\sigma = \sqrt{\frac{1}{N} \sum_i (v_i - \mu)},
    \end{aligned}
\end{equation}

where $v_i$ is an element of a noise-free tensor $v$, and $v$ represents the input/weight/inter-layer tensor. Next, $\mu$ is the estimated mean of $v$, and $\eta$ is the additive noise term, which is a random variable following the Gaussian distribution. Finally, $\alpha_i$ is the coefficient which scales the magnitude of the injected noise, and it is a learnable parameter which is optimized for the network's robustness. 
The default setting in~\cite{Rakin2019ParametricNI} is to apply PNI to weight tensors of convolutional and fully-connected layers (denoted as PNI-W) , and to share the element-wise noise coefficient $\alpha_i$ for all elements of a specific weight tensor (denoted as layerwise). We also evaluate the setting where the PNI is applied to tensors which are outputs of the convolutional and fully connected layers (denoted as PNI-A-a), because it has also shown good results~\cite{Rakin2019ParametricNI}. Furthermore, we also explore sharing $\alpha_i$ just for different channels inside the tensor (denoted as channelwise), or having different $\alpha_i$ for different elements (denoted as elementwise).

\section{Experiments}
\label{sec:experiments}

\subsection{Experimental setup}
\noindent\textbf{Adversarial attack strategies.}
In general, adversarial attacks exploit the differentiability of the network $f(x)$ and its prediction loss $\mathcal{L}(f(x),l)$, with respect to the input image $x$, where $l$ is the label. 
The attack aims to slightly modify the input $x$ to maximize the prediction loss for the correct label $l$. To craft stronger adversarial samples, the fast gradient sign method (FGSM)~\cite{Goodfellow2015ExplainingAH} is repeated K times with a step size of $\alpha$, followed by a projection to an $\epsilon$ hypercube around the initial sample $x$, $\hat{x}^{k} = \Pi_{x, \epsilon} \left[\hat{x}^{k-1} + \alpha \, \text{sgn}(\nabla_x \mathcal{L}(f(\hat{x}^{k-1}), l))\right]$. This is known as the projected gradient descent (PGD-K) attack~\cite{Madry2018TowardsDL}.
Furthermore, the expectation-over-transformation (EOT) gradient estimation is usually more effective when dealing with noise inside the network, because of common gradient obfuscations~\cite{Athalye2018ObfuscatedGG}. This can be viewed as using PGD~\cite{Madry2018TowardsDL}
with the proxy gradient, $\mathbb{E}_{q(z)} \left[\nabla_x \mathcal{L}(f(\hat{x}^{k-1}, z), l)\right] \approx \frac{1}{T} \sum_{t=1}^{T} \nabla_x \mathcal{L}(f(\hat{x}^{k-1}, z_t), l)$ ,
where $q(z)$ represents the distribution of the noise ${z\sim q(z)}$ injected into the randomized classifier $f(x,z)$.

\noindent\textbf{Adversarial vulnerability metrics.}
In order to evaluate adversarial robustness, we craft adversarial examples for each image in the validation set with the aforementioned attacks, and test the model’s accuracy on those adversarial examples.
When crafting each adversarial example, we initialize the attack's starting point randomly, inside the $\epsilon-$hypercube centered on $x$.
We restart this procedure R times to find the strongest attack and always set the step size $\alpha=1$.

\noindent\textbf{Dataset.} For conducting experiments, we use the ILSVRC-2012 ImageNet dataset, containing 1.2M training and 50000 validation images grouped into 1000 classes~\cite{Russakovsky2015ImageNet}. 

\noindent\textbf{Adversarial training.}
PNI is trained with the help of adversarial training~\cite{Rakin2019ParametricNI}.
Since we use the more computationally demanding ImageNet dataset, we employ the recent efficient adversarial training procedure described in~\cite{Wong2020FastIB}.
Following~\cite{Wong2020FastIB}, the step sizes during adversarial training are $\alpha= \{\frac{2.5}{255}, \frac{5}{255} \}$ for $\epsilon= \{ \frac{2}{255}, \frac{4}{255}\}$, respectively. The models are evaluated for the same $\epsilon$ used during the training.

\noindent\textbf{Baselines.} 
For all experiments, the baseline is the original network, without the PNI stochasticity. In a way, this baseline serves as an ablated model. A significant improvement over this baseline is necessary to claim any improvement in robustness. More importantly, the same must hold even with EoT gradient estimation so as to ensure that the robustness is not mainly due to gradient obfuscation.  

\noindent\textbf{Implementation details.}
For the main experiments, we use the ResNet architecture, where every convolutional layer has been extended with the PNI, as described in~\eqref{eq:PNI}. More specifically, we use the ResNet-50, as it provides a good trade-off between performance and computational complexity, and we train it from scratch. We use 100 randomly selected classes, because of the high computational demand of the ImageNet dataset, adversarial training, and adversarial evaluation altogether. During training, we use hyperparameters recently proposed in~\cite{Touvron21aDeIT}, which have shown to work well for ResNets~\cite{Wightman2021ResnetStrikes}. We train for 150, because it turns out to be sufficient in this setting with 100 classes.

Furthermore, we also preform experiments on the DeiT-S transformer~\cite{Touvron21aDeIT}, since transformer architectures are becoming very popular and relevant in computer vision. 
The DeiT-S has the parameter count and computational complexity similar to a ResNet-50. 
We extend the fully-connected layer, just after the activation (in the MLP block), with the PNI on its weights (PNI-W-fc2). 
This experiment also analyzes the case of less aggressive noise, since PNI is not used in all parametrized blocks of the transformer.
The initial experiments, like described in the case of ResNet, did not perform as well, probably because of the data hungry nature of transformers.
Therefore, we use the whole ImageNet dataset for this experiment.
However, because of high computational demand, we start from pre-trained models on ImageNet, like the ones described in~\cite{Touvron21aDeIT}. During the fine-tuning, which lasts for $20$ epochs, we use the AdamW optimizer and a cosine scheduler with a learning rate of $10^{-5}$, which gradually decays to $10^{-6}$.

During every evaluation, we restart the attack 5 times to construct a stronger attack and we use 10 steps for the PGD attack (PGD-10). The number of samples for the EoT estimation is 25 in the case of ResNet50 on 100 classes (EoT-25), and 5 in the case of DeiT-S on all 1000 classes (EoT-5).

\subsection{Results}
\label{ssec:results}
\begin{table}[t]
\centering
\tableFont
\captionsetup{font=small}
\caption{\textbf{Robustness of models with and without PNI}. 
Inserting various forms of PNI improves the adversarial robustness of both the ResNet50 and DeiT-S over the respective baselines, for both FGSM and PGD-10 attacks. In contrast, when EoT is used to estimate the gradients during the attacks, the effect of PNI is even detrimental, weakening the desired robustness. This is a clear sign of gradient obfuscation being the main source of robustness.
}


\begin{subtable}{1\columnwidth}
\centering
\caption{ResNet50 trained from scratch with PNI, on 100 ImageNet classes.}
\begin{tabular}{|c||c|c|c|c|c|}

\hline
& \multicolumn{5}{c|}{\makecell{Accuracy $\uparrow$}} \\ \cline{2-6}

& \makecell{Clean \\ samples}
& {\makecell{FGSM \\ attack}} 
& {\makecell{FGSM \\ attack \\ (EoT$-25$)}} 
& {\makecell{PDG$-10$ \\ attack}} 
& {\makecell{PDG$-10$ \\ attack \\ (EoT$-25$)}} 
\\  \hline \hline 

& \multicolumn{5}{c|}{\makecell{Adversarial training from scratch with $\epsilon=\frac{2}{255}$}} 
\\ \hline \hline

ResNet50
& $82.90\%$ & \multicolumn{2}{c|}{$72.24\%$} & \multicolumn{2}{c|}{$65.44\%$} \\ \hline 

\makecell{ResNet50 + PNI-W  \\ (layerwise)}
& $83.16\%$ & $77.60\% \greenA $ & $71.30\% \redA$  & $70.26\% \greenA$ & $62.70\% \redA$ \\ \hline

\makecell{ResNet50 + PNI-W \\(channelwise)}
& $84.50\%$ & $77.62\% \greenA $ & $70.20\% \redA$  & $68.92\% \greenA$ & $60.02\% \redA$ \\ \hline

\makecell{ResNet50 + PNI-W \\(elementwise)}
& $83.18\%$ & $76.00\% \greenA $ & $69.96\% \redA$  & $68.74\% \greenA$ & $61.70\% \redA$ \\ \hline

\makecell{ResNet50 + PNI-A-a \\(layerwise)}
& $85.06\%$ & $75.52\% \greenA $ & $69.16\% \redA$  & $66.74\% \greenA$ & $59.04\% \redA$ \\ \hline 

\makecell{ResNet50 + PNI-A-a  \\(channelwise)}
& $85.20\%$ & $75.38\% \greenA $ & $69.12\% \redA$  &  $66.64\% \greenA$ & $59.08\% \redA$ \\ \hline \hline

& \multicolumn{5}{c|}{\makecell{Adversarial training from scratch with $\epsilon=\frac{4}{255}$}}
\\\hline \hline


Res50 
& $78.12\%$ & \multicolumn{2}{c|}{$61.72\%$} & \multicolumn{2}{c|}{$50.94\%$} \\ \hline 

\makecell{ResNet50 + PNI-W \\(layerwise)}
& $82.02\%$  & $71.24\% \greenA $ & $60.64\% \redA$  & $60.60\% \greenA$ & $44.50\% \redA$ \\ \hline

\makecell{ResNet50 + PNI-W \\(channelwise)}
& $82.54\%$  & $72.68\% \greenA $ & $60.36\% \redA$  & $60.38\% \greenA$ & $42.62\% \redA$ \\ \hline

\makecell{ResNet50 + PNI-W \\(elementwise)}
& $79.76\%$ & $72.42\% \greenA $ & $62.86\% \greenA$  & $63.40\% \greenA$ & $48.00\% \redA$ \\ \hline

\makecell{ResNet50 + PNI-A-a \\(layerwise)}
& $82.08\%$ & $67.32\% \greenA $ & $56.90\% \redA$   & $55.12\% \greenA$ & $42.64\% \redA$ \\ \hline 

\makecell{ResNet50 + PNI-A-a \\(channelwise)}
& $81.92\%$  & $67.90\% \greenA $ & $57.84\% \redA$  & $55.92\% \greenA$ & $43.34\% \redA$ \\ \hline
 

\end{tabular}
\label{tab:PNI_merged_resnet}
\end{subtable}

\begin{subtable}{1\columnwidth}
\caption{DeiT-S fine-tuned with PNI, on all 1000 ImageNet classes.}
\centering
\begin{tabular}{|c||c|c|c|c|c|}

\hline
& \multicolumn{5}{c|}{\makecell{Accuracy $\uparrow$}} \\ \cline{2-6}
 
& \makecell{Clean \\ samples}
& {\makecell{FGSM \\ attack}} 
& {\makecell{FGSM \\ attack \\ (EoT$-5$)}} 
& {\makecell{PDG$-10$ \\ attack}} 
& {\makecell{PDG$-10$ \\ attack \\ (EoT$-5$)}} 
\\  \hline \hline 
  
& \multicolumn{5}{c|}{\makecell{Adversarial fine-tuning with $\epsilon=\frac{2}{255}$}} \\ \hline \hline

DeiT-S 
& $71.61\%$ & \multicolumn{2}{c|}{$53.27\%$} & \multicolumn{2}{c|}{$42.33\%$} \\ \hline 

\makecell{ DeiT-S + PNI-W-fc2 \\(layerwise)}
& $73.50\%$ & $58.06\% \greenA$ & $54.40\% \greenA$ & $44.84\% \greenA$ & $40.60\% \redA$ \\ \hline

\makecell{ DeiT-S + PNI-W-fc2 \\(channelwise)}

& $72.88\%$ & $57.02\% \greenA$ & $53.58\% \greenA$ & $44.47\% \greenA$ & $40.96\% \redA$ \\ \hline
 
\makecell{ DeiT-S + PNI-W-fc2 \\(elementwise)}
& $72.51\%$ & $56.66\% \greenA$ & $53.49\% \greenA$ & $44.49\% \greenA$ & $41.08\% \redA$ \\ \hline \hline

& \multicolumn{5}{c|}{\makecell{Adversarial fine-tuning with $\epsilon=\frac{4}{255}$}} \\ \hline \hline

DeiT-S 
& $65.04\%$ & \multicolumn{2}{c|}{$39.98\%$} & \multicolumn{2}{c|}{$27.24\%$} \\ \hline 

\makecell{ DeiT-S + PNI-W-fc2 \\(layerwise)}
& $68.77\%$ & $47.12\% \greenA$ & $41.56\% \greenA$ & $31.07\% \greenA$ & $25.90\% \redA$ \\ \hline

\makecell{ DeiT-S + PNI-W-fc2 \\(channelwise)}
& $67.40\%$ & $45.07\% \greenA$ & $40.78\% \greenA$ & $30.22\% \greenA$ & $26.11\% \redA$ \\ \hline
 
\makecell{ DeiT-S + PNI-W-fc2 \\(elementwise)}
& $67.02\%$ & $44.24\% \greenA$ & $40.61\% \greenA$ & $29.99\% \greenA$ & $26.13\% \redA$ \\ \hline

\end{tabular}
\label{tab:PNI_merged_trf}
\end{subtable}

\label{tab:PNI_merged}
\end{table}

In Table~\ref{tab:PNI_merged} we see that inserting various forms of PNI improves the adversarial robustness of both the ResNet50 and DeiT-S over the respective baselines, for both FGSM and PGD-10 attacks. In contrast, when EoT is used to estimate the gradients during the attacks, the effect of PNI is even detrimental, weakening the desired robustness. Note that being effective against regular PGD, but ineffective against PGD with EoT, is clear evidence for gradient obfuscation being the main source of robustness. Rather than strengthening the robustness of the visual features, such defenses rather make it harder to find the adversarial example with gradient-based attacks. EoT however allows to uncover the adversarial direction by averaging multiple noisy gradient samples, and exposes the original vulnerability of the network. Note that the results of Table~\ref{tab:PNI_merged} (a) and (b) cannot be directly compared, due to the differences in the following aspects: number of classes, number of EoT samples, network backbones, and the training protocols. Nevertheless, both (a) and (b) support our conclusions.



\section{Conclusion}
\label{sec:conclusion}
In this paper, we reflect on the problem of gradient obfuscation in the case of stochastic defense techniques against adversarial attacks.
Athalye et al.~\cite{Athalye2018ObfuscatedGG} observed five common characteristics, when the improvement in robustness is mainly caused by gradient obfuscation. They also stated that ``these behaviors may not perfectly characterize all cases of masked gradients". Despite this, it has become a trend to claim that obfuscated gradients are not the main source of improvements in robustness, if none of these five characteristics hold true\trendCiteShort. We refute such claims on a large-scale dataset by providing a counterexample. 

In particular, we have shown that the popular Parametric Noise Injection (PNI) exploits gradient obfuscation to improve robustness, despite of passing the five characteristics checklist test.  The exploitation of the gradient obfuscation is unveiled based on the following observations:
\begin{itemize}[topsep=0pt,itemsep=-1ex,partopsep=1ex,parsep=1ex]
\item PNI passes the five characteristics checklist test~\cite{Rakin2019ParametricNI}.
\item Adding PNI improves the adversarial robustness towards FGSM and PGD attacks.
\item Adding PNI is detrimental for robustness towards attacks using gradients estimated with EoT.
\end{itemize}
%
\noindent This counterexample allows us to conclude that the gradient obfuscation checklist test is not sufficient to determine whether or not the gradient obfuscation is the main source of robustness improvements. Therefore, only using the gradient obfuscation checklist test  gives us a false sense of security.
Needless to say, the provided counterexample is sufficient to make the above conclusion. Henceforth, we recommend to  include EoT-based attacks in the gradient obfuscation test. Please note that even with the EoT criterion, not all cases of obfuscated gradients may be perfectly covered.









{\small
\bibliographystyle{ieee_fullname}
\bibliography{ms}

\begin{thebibliography}{10}\itemsep=-1pt

\bibitem{Addepalli_2021_CVPR}
Sravanti Addepalli, Samyak Jain, Gaurang Sriramanan, and R.~Venkatesh Babu.
\newblock Boosting adversarial robustness using feature level stochastic
  smoothing.
\newblock In {\em CVPR Workshops}, June 2021.

\bibitem{Athalye2018ObfuscatedGG}
Anish Athalye, Nicholas Carlini, and David~A. Wagner.
\newblock Obfuscated gradients give a false sense of security: Circumventing
  defenses to adversarial examples.
\newblock In {\em ICML}, 2018.

\bibitem{Athalye2018SynthesizingRA}
Anish Athalye, Logan Engstrom, Andrew Ilyas, and Kevin Kwok.
\newblock Synthesizing robust adversarial examples.
\newblock {\em ArXiv}, 2018.

\bibitem{Ciss2017ParsevalNI}
Moustapha Ciss{\'e}, Piotr Bojanowski, Edouard Grave, Yann Dauphin, and Nicolas
  Usunier.
\newblock Parseval networks: Improving robustness to adversarial examples.
\newblock {\em ArXiv}, 2017.

\bibitem{Cohen2019CertifiedAR}
Jeremy~M. Cohen, Elan Rosenfeld, and J.~Zico Kolter.
\newblock Certified adversarial robustness via randomized smoothing.
\newblock In {\em ICML}, 2019.

\bibitem{Croce2019ProvableRO}
Francesco Croce, Maksym Andriushchenko, and Matthias Hein.
\newblock Provable robustness of relu networks via maximization of linear
  regions.
\newblock {\em ArXiv}, 2019.

\bibitem{Eustratiadis2021WeightCovAlign}
Panagiotis Eustratiadis, Henry Gouk, Da Li, and Timothy Hospedales.
\newblock Weight-covariance alignment for adversarially robust neural networks.
\newblock In {\em ICML}, 2021.

\bibitem{Finlay2019ScaleableIG}
Chris Finlay and Adam~M. Oberman.
\newblock Scaleable input gradient regularization for adversarial robustness.
\newblock {\em ArXiv}, 2019.

\bibitem{Goodfellow2015ExplainingAH}
Ian~J. Goodfellow, Jonathon Shlens, and Christian Szegedy.
\newblock Explaining and harnessing adversarial examples.
\newblock {\em CoRR}, 2015.

\bibitem{Gu2015TowardsDN}
Shixiang~Shane Gu and Luca Rigazio.
\newblock Towards deep neural network architectures robust to adversarial
  examples.
\newblock {\em CoRR}, 2015.

\bibitem{Jakubovitz2018ImprovingDR}
Daniel Jakubovitz and Raja Giryes.
\newblock Improving dnn robustness to adversarial attacks using jacobian
  regularization.
\newblock {\em ArXiv}, 2018.

\bibitem{Jeddi_2020_CVPR}
Ahmadreza Jeddi, Mohammad~Javad Shafiee, Michelle Karg, Christian
  Scharfenberger, and Alexander Wong.
\newblock Learn2perturb: An end-to-end feature perturbation learning to improve
  adversarial robustness.
\newblock In {\em CVPR}, June 2020.

\bibitem{Kundu2021HIRE_SNN}
Souvik Kundu, Massoud Pedram, and Peter~A. Beerel.
\newblock Hire-snn: Harnessing the inherent robustness of energy-efficient deep
  spiking neural networks by training with crafted input noise.
\newblock In {\em ICCV}, October 2021.

\bibitem{Kurakin2017AdversarialML}
Alexey Kurakin, Ian~J. Goodfellow, and Samy Bengio.
\newblock Adversarial machine learning at scale.
\newblock {\em ArXiv}, 2017.

\bibitem{lecun2015deep}
Yann LeCun, Yoshua Bengio, and Geoffrey Hinton.
\newblock Deep learning.
\newblock {\em nature}, 521, 2015.

\bibitem{Lee2021GradDivAR}
Sungyoon Lee, Hoki Kim, and Jaewook Lee.
\newblock Graddiv: Adversarial robustness of randomized neural networks via
  gradient diversity regularization.
\newblock {\em ArXiv}, 2021.

\bibitem{Lecuyer2019CertifiedDifferential}
Mathias Lécuyer, Vaggelis Atlidakis, Roxana Geambasu, and Daniel Hsu.
\newblock Certified robustness to adversarial examples with differential
  privacy.
\newblock 2019.

\bibitem{Madry2018TowardsDL}
Aleksander Madry, Aleksandar Makelov, Ludwig Schmidt, Dimitris Tsipras, and
  Adrian Vladu.
\newblock Towards deep learning models resistant to adversarial attacks.
\newblock {\em ArXiv}, 2018.

\bibitem{mihajlovic2018CommonAdv}
Marko Mihajlović and Nikola Popović.
\newblock Fooling a neural network with common adversarial noise.
\newblock In {\em 2018 19th IEEE Mediterranean Electrotechnical Conference
  (MELECON)}, 2018.

\bibitem{Rakin2019ParametricNI}
Adnan~Siraj Rakin, Zhezhi He, and Deliang Fan.
\newblock Parametric noise injection: Trainable randomness to improve deep
  neural network robustness against adversarial attack.
\newblock {\em CVPR}, 2019.

\bibitem{Ross2018ImprovingTA}
Andrew~Slavin Ross and Finale Doshi-Velez.
\newblock Improving the adversarial robustness and interpretability of deep
  neural networks by regularizing their input gradients.
\newblock In {\em AAAI}, 2018.

\bibitem{Russakovsky2015ImageNet}
Olga Russakovsky, Jia Deng, Hao Su, Jonathan Krause, Sanjeev Satheesh, Sean Ma,
  Zhiheng Huang, Andrej Karpathy, Aditya Khosla, Michael Bernstein,
  Alexander~C. Berg, and Li Fei-Fei.
\newblock {ImageNet Large Scale Visual Recognition Challenge}.
\newblock {\em IJCV}, 2015.

\bibitem{Salman2019ProvablyRD}
Hadi Salman, Greg Yang, Jungshian Li, Pengchuan Zhang, Huan Zhang, Ilya~P.
  Razenshteyn, and S{\'e}bastien Bubeck.
\newblock Provably robust deep learning via adversarially trained smoothed
  classifiers.
\newblock In {\em NeurIPS}, 2019.

\bibitem{Szegedy2014IntriguingPO}
Christian Szegedy, Wojciech Zaremba, Ilya Sutskever, Joan Bruna, D. Erhan,
  Ian~J. Goodfellow, and Rob Fergus.
\newblock Intriguing properties of neural networks.
\newblock {\em CoRR}, 2014.

\bibitem{Touvron21aDeIT}
Hugo Touvron, Matthieu Cord, Matthijs Douze, Francisco Massa, Alexandre
  Sablayrolles, and Herve Jegou.
\newblock Training data-efficient image transformers \& distillation through
  attention.
\newblock In {\em ICML}, July 2021.

\bibitem{Wightman2021ResnetStrikes}
Ross Wightman, Hugo Touvron, and Hervé Jégou.
\newblock Resnet strikes back: An improved training procedure in timm.
\newblock {\em ArXiv}, 2021.

\bibitem{Wong2020FastIB}
Eric Wong, Leslie Rice, and J.~Zico Kolter.
\newblock Fast is better than free: Revisiting adversarial training.
\newblock {\em ArXiv}, 2020.

\bibitem{Yang2022FeatureUncrt}
Hao Yang, Min Wang, Zhengfei Yu, and Yun Zhou.
\newblock Rethinking feature uncertainty in stochastic neural networks for
  adversarial robustness.
\newblock {\em CoRR}, 2022.

\bibitem{Zhang2019TheoreticallyPT}
Hongyang~R. Zhang, Yaodong Yu, Jiantao Jiao, Eric~P. Xing, Laurent~El Ghaoui,
  and Michael~I. Jordan.
\newblock Theoretically principled trade-off between robustness and accuracy.
\newblock In {\em ICML}, 2019.

\end{thebibliography}
}

\clearpage

\end{document}